\documentclass{article}

\usepackage{PRIMEarxiv}
\usepackage[numbers]{natbib}
\usepackage[utf8]{inputenc} 
\usepackage[T1]{fontenc}    
\usepackage{hyperref}       
\usepackage{url}            
\usepackage{booktabs}       
\usepackage{amsfonts}       
\usepackage{nicefrac}       
\usepackage{microtype}      
\usepackage{lipsum}
\usepackage{fancyhdr}       
\usepackage{graphicx}       
\usepackage{times}  
\usepackage{helvet}  
\usepackage{courier}  
\usepackage{algorithmic}
\usepackage{xcolor}
\usepackage{hyperref}
\usepackage{enumitem}
\usepackage{tabularx}
\usepackage{threeparttable}
\newcommand{\scifig}{{\sc ACL-Fig}}
\newcommand{\scifigp}{{\sc ACL-Fig-pilot}}
\graphicspath{{media/}}     

\title{ACL-Fig: A Dataset for Scientific Figure Classification
}

\author{
  Zeba Karishma, Shaurya Rohatgi, Kavya Shrinivas Puranik \\
  The Pennsylvania State University \\
  \texttt{zebakarishma@gmail.com, \{szr207, kzp5555\}@psu.edu} \\
  \And
  Jian Wu \\
  Old Dominion University \\
  \texttt{jwu@cs.odu.edu} \\
   \And
  C. Lee Giles \\
  The Pennsylvania State University \\
  \texttt{clg20@psu.edu} \\
}

\begin{document}
\maketitle

\begin{abstract}
Most existing large-scale academic search engines are built to retrieve text-based information. However, there are no large-scale retrieval services for scientific figures and tables. One challenge for such services is understanding scientific figures' semantics, such as their types and purposes. A key obstacle is the need for datasets containing annotated scientific figures and tables, which can then be used for classification, question-answering, and auto-captioning. Here, we develop a pipeline that extracts figures and tables from the scientific literature and a deep-learning-based framework that classifies scientific figures using visual features. Using this pipeline, we built the first large-scale automatically annotated corpus, \scifig\, consisting of 112,052 scientific figures extracted from $\approx56$K research papers in the ACL Anthology. The \scifigp\ dataset contains 1,671 manually labeled scientific figures belonging to 19 categories. The dataset is  accessible at 
\href{https://huggingface.co/datasets/citeseerx/ACL-fig}{https://huggingface.co/datasets/citeseerx/ACL-fig}
under a CC BY-NC license.
\end{abstract}

\section{Introduction}

Figures are ubiquitous in scientific papers  illustrating experimental and analytical results. We refer to these figures as \emph{scientific figures} to distinguish them from natural images, which usually contain richer colors and gradients. Scientific figures provide a compact way to present numerical and categorical data, often facilitating researchers in drawing insights and conclusions. Machine understanding of scientific figures can assist in developing effective retrieval systems from the hundreds of millions of scientific papers readily available on the Web \cite{khabsa2014plosone}.
The state-of-the-art machine learning models can parse captions and shallow semantics for specific categories of scientific figures. \cite{Siegel_2018} However, the task of reliably classifying general scientific figures based on their visual features remains a challenge.
\begin{figure}
     \centering
     \includegraphics[width = {0.6\textwidth}]{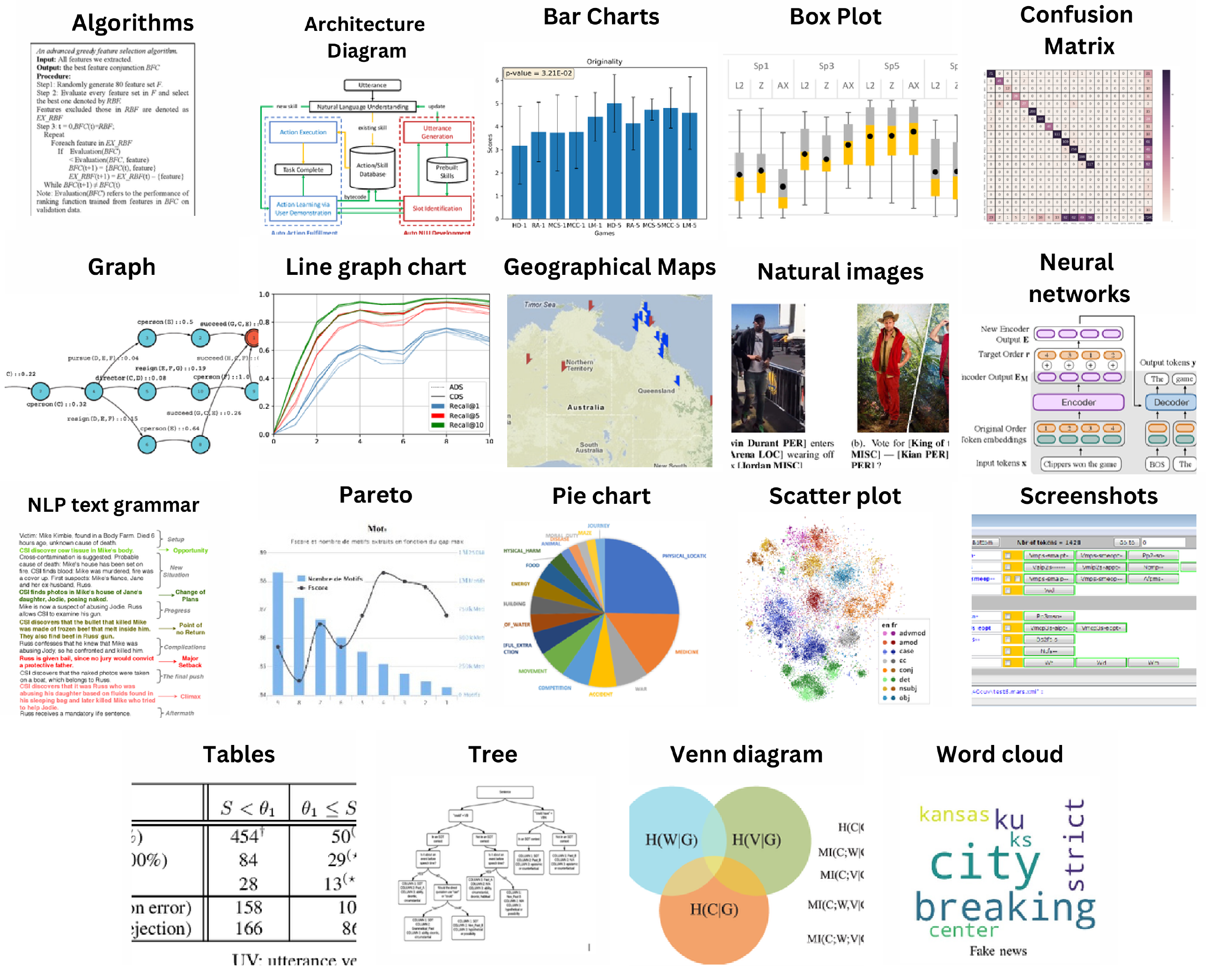}
     \caption{Example figures of each type in {\scifigp}.}
     \label{fig:samImg}
 \end{figure}
 
Here, we propose a pipeline to build categorized and contextualized scientific figure datasets. Applying the pipeline on 55,760 papers in the ACL Anthology (downloaded from https://aclanthology.org/ in mid-2021), we built two datasets: \scifig\ and \scifigp. \scifig\ consists of 112,052 scientific figures, their captions, inline references, and  metadata. \scifigp\ (Figure \ref{fig:samImg}) is a subset of unlabeled \scifig, consisting of 1671 scientific figures, which were manually labeled into 19 categories. The \scifigp\ dataset was used as a benchmark for scientific figure classification. The pipeline is open-source and configurable, enabling others to expand the datasets  from other scholarly datasets with pre-defined or new labels. 

\section{Related Work}

\paragraph{Scientific Figures Extraction}
Automatically extracting figures from scientific papers is essential for many downstream tasks, and many frameworks have been developed. 
A multi-entity extraction framework called PDFMEF incorporating a figure extraction module was proposed \cite{Wu2015PDFMEFAM}. Shared tasks such as ImageCLEF \cite{Herrera2015OverviewOT} drew attention to compound figure detection 
and separation. 
\cite{Clark2015LookingBT} proposed a framework called {\sc PDFFigures} that extracted figures and captions in research papers. The authors extended their work and built a more robust framework called {\sc PDFFigures2} \cite{Clark2016PDFFigures2M}. {\sc DeepFigures} was later proposed to incorporate deep neural network models \cite{Siegel_2018}. 

\paragraph{Scientific Figure Classification}
Scientific figure classification \cite{Savva2011ReVisionAC, Choudhury2015AnAF} aids machines in understanding figures. Early work used a visual bag-of-words representation with a support vector machine classifier \cite{Savva2011ReVisionAC}. \cite{Zhou2000HoughTF} applied hough transforms to recognize bar charts in document images. 
\cite {Siegel2016FigureSeerPR} used handcrafted features to classify charts in scientific documents. 
\cite{tange2016deepchart} combined convolutional neural networks (CNNs) and the deep belief networks, which showed improved performance compared with feature-based classifiers . 


\begin{table}
\centering
\begin{threeparttable}[b]
\small
\centering
\caption{Scientific figure classification datasets. }
\label{Ch1-table:Stats}
\begin{tabular}{l|l|l|l} 
\toprule
Dataset                    & \textbf{Labels} & \textbf{\#Figures} & \textbf{Image Source}       \\ 
\midrule
Deepchart                  & 5               & 5,000               & Web Image                   \\
Figureseer\tnote{1}                 & 5               & 30,600             & Web Image                   \\
Prasad et al.              & 5               & 653                & Web Image                   \\
Revision                   & 10              & 2,000               & Web Image                   \\
FigureQA\tnote{3}                  & 5               & 100,000             & Synthetic figures           \\ 
\midrule
DeepFigures & 2 & 1,718,000 & Scientific Papers\\
DocFigure\tnote{2}                  & 28              & 33,000             & Scientific Papers           \\
\textbf{\scifigp}      & \textbf{19}     & \textbf{1,671}      & Scientific Papers  \\
{\scifig} (inferred)\tnote{4} & -               & \textbf{112,052}    & Scientific Papers \\
\bottomrule
\end{tabular}
\begin{tablenotes}
\item [1] {Only 1000 images are public}.
\item [2] {Not publicly available.}
\item [3] {Scientific-style synthesized data}.
\item [4] {\scifig}-inferred does not contain human-assigned labels.
\end{tablenotes}
\end{threeparttable}
\end{table}

\paragraph{Figure classification Datasets}
There are several existing datasets for figure classification such as DocFigure \citep{8892905}, FigureSeer \cite{Siegel2016FigureSeerPR}, Revision \cite{Savva2011ReVisionAC}, and datasets presented by \cite{Karthikeyani2012MachineLC} (Table~\ref{Ch1-table:Stats}). 
FigureQA is a public dataset that is similar to ours, consisting of over one million question-answer pairs grounded in over 100,000 synthesized scientific images \cite{kahou2018figureqa} with five styles.
Our dataset is different from FigureQA because the figures were directly extracted from research papers. Especially, the training data of {\sc DeepFigures} are from arXiv and PubMed, labeled with only ``figure'' and ``table'', and does not include fine-granular labels. Our dataset contains fine-granular labels, inline context, and is compiled from a different domain.  


\begin{figure}[htb]
    \centering
    \includegraphics[width={0.5\textwidth}]{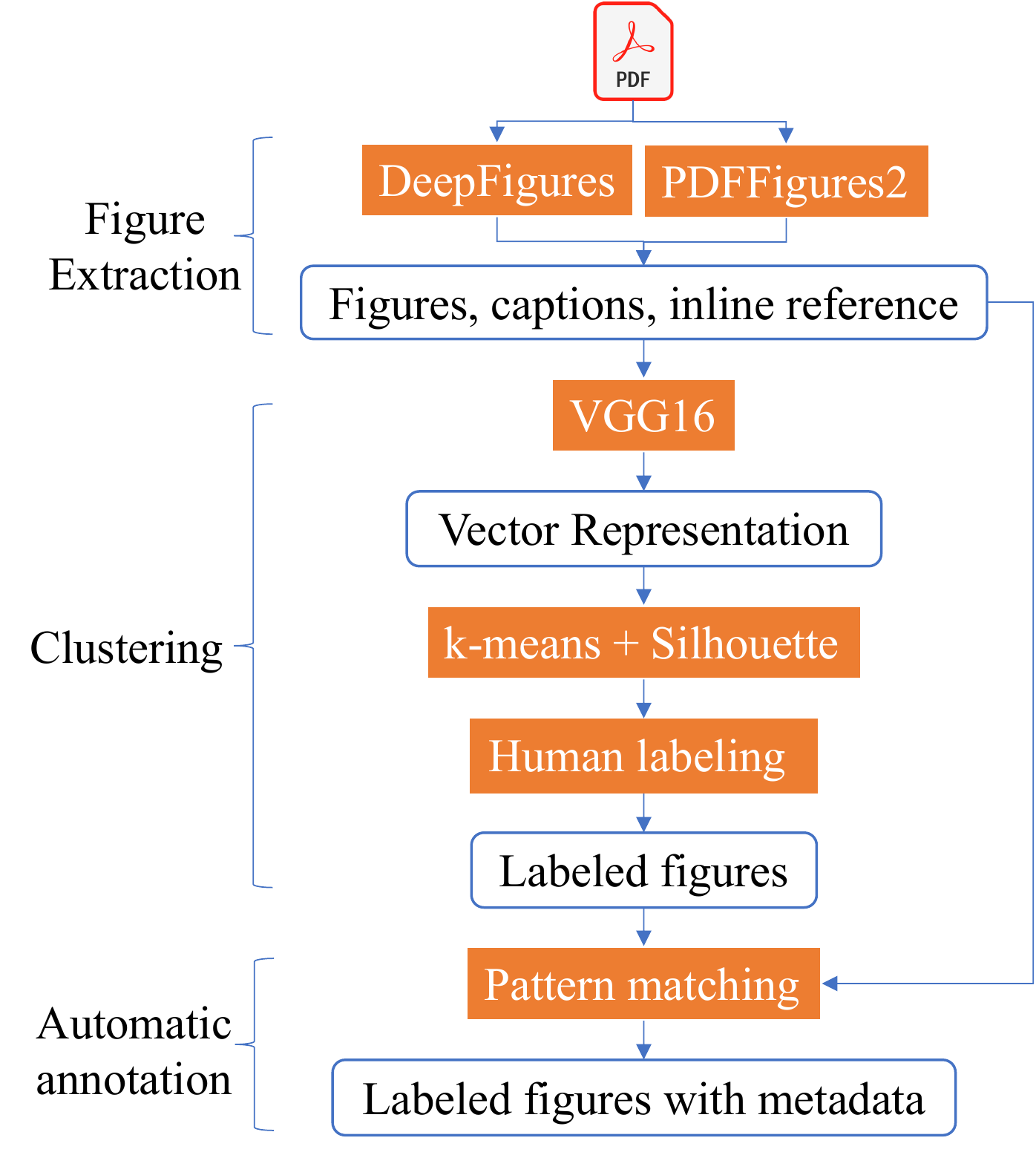}
    \caption{Overview of the data generation pipeline.}\label{fig:arch}
\end{figure}

\section{Data Mining Methodology}
The ACL Anthology is a sizable, well-maintained PDF corpus with clean metadata covering papers in computational linguistics with freely available full-text. Previous work on figure classification used a set of pre-defined categories (e.g., \cite{kahou2018figureqa}, which may only cover some figure types.  We use an unsupervised method to determine figure categories to overcome this limitation. After the category label is assigned, each figure is automatically annotated with metadata, captions, and inline references. The pipeline includes 3 steps: figure extraction,  clustering, and automatic annotation (Figure~\ref{fig:arch}).

\subsection{Figure Extraction}
To mitigate the potential bias of a single figure extractor, we extracted figures using {\sc pdffigures2} \cite{Clark2016PDFFigures2M}  and {\sc deepfigures} \cite{Siegel_2018} which work in different ways. {\sc PDFFigures2} first identifies  captions and the body text because they are identified relatively accurately. Regions containing figures can then be located by identifying rectangular bounding boxes adjacent to captions that do not overlap with the body text. {\sc DeepFigures}  uses the distant supervised learning method to induce labels of figures from a large collection of scientific documents in LaTeX and XML format. The model is based on TensorBox, applying the Overfeat detection architecture to image embeddings generated using ResNet-101 \cite{Siegel_2018}. We utilized the publicly available model weights\footnote{https://github.com/allenai/deepfigures-open} trained on 4M induced figures and 1M induced tables for extraction. The model outputs the bounding boxes of figures and tables. Unless otherwise stated, we collectively refer to figures and tables together as ``figures''. 
We used multi-processing to process PDFs. Each process extracts figures following the steps below. The system processed, on average, 200 papers per minute on a Linux server with 24 cores.
\begin{enumerate}[leftmargin=*,noitemsep]
    \item Retrieve a paper identifier from the job queue.
    \item Pull the paper from the file system.
    \item Extract figures and captions from the paper.
    \item Crop the figures out of the rendered PDFs using detected bounding boxes.
    \item Save cropped figures in PNG format and the metadata in JSON format. 
\end{enumerate}

\subsection{Clustering Methods\label{sec:clusteringmethod}} 
Next, we use an unsupervised method to label
extracted figures automatically.  We extract visual features using
VGG16 \cite{simonyan2014very}, pretrained on ImageNet \cite{deng2009imagenet}. 
All input figures are scaled to a dimension of $224\times224$ to be compatible with the input requirement of VGG16. The features were extracted from the second last hidden (dense) layer, consisting of
4096 features. Principal Component Analysis was adopted to reduce the dimension to 1000.

Next, we cluster figures represented by the 1000-dimension vectors using $k$-means clustering. We compare two heuristic methods to determine the optimal number of clusters, including the Elbow method 
and the Silhouette Analysis \citep{rousseeue1987silhouettes}. The Elbow method examines the \emph{explained variation}, a measure that quantifies the difference between the between-group variance to the total variance,  as a function of the number of clusters. The pivot point (elbow) of the curve determines the number of clusters. 

Silhouette Analysis determines the number of clusters by measuring the distance between clusters. 
It considers multiple factors such as variance, skewness, and high-low differences and is usually preferred to the Elbow method. The Silhouette plot displays how close each point in one cluster is to points in the neighboring clusters, allowing us to assess the cluster number visually.  

\subsection{Linking Figures to Metadata}
This module associates figures to metadata, including captions, inline reference, figure type, figure boundary coordinates, caption boundary coordinates, and figure text (text appearing on figures, only available for results from {\sc PDFFigures2}). The figure type is determined in the clustering step above. The inline references are obtained using GROBID (see below). The other metadata fields were output by figure extractors. {\sc PDFFigures2} and {\sc DeepFigures} extract the same metadata fields except for ``image text'' and ``regionless captions'' (captions for which no figure regions were found), which are only available for results of {\sc PDFFigures2}.


An inline reference is a text span that contains a reference to a figure or a table. Inline references can help to understand the relationship between text and the objects it refers to. After processing a paper, GROBID outputs a TEI file (a type of XML file), containing marked-up full-text and references. We locate inline references using regular expressions and extract the sentences containing reference marks. 

\section{Results}
\subsection{Figure Extraction}


\begin{figure}[h!]
     \centering
     \includegraphics[width = {0.75\textwidth}]{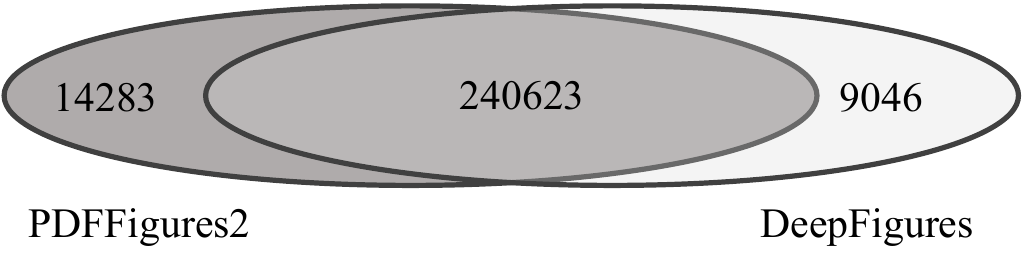}
     \caption{\label{fig:venndiagram}Numbers of extracted images.}
 \end{figure}
The numbers of  figures extracted by {\sc PDFFigures2} and {\sc DeepFigures} are illustrated in Figure~\ref{fig:venndiagram}, which indicates a significant overlap between figures extracted by two software packages. However, either package extracted ($\approx5\%$) figures that were not extracted by the other package. By inspecting a random sample of figures extracted by either software package, we found that {\sc DeepFigures} tended to miss cases in which two figures were vertically adjacent to each other. 
We took the union of all figures extracted by both software packages to build the {\scifig}  dataset, which contains a total of 263,952 figures. All images extracted are converted to 100 DPI using standard OpenCV libraries. The total size of the data is $\sim25$GB before compression. Inline references were extracted using GROBID. About 78\% figures have inline references. 



\subsection{Automatic Figure Annotation}
The extraction outputs 151,900 tables and 112,052 figures. Only the figures were clustered using the $k$-means algorithm. We varied $k$ from 2 to 20 with an increment of 1 to determine the number of clusters. The results were analyzed using the Elbow method and Silhouette Analysis. No evident elbow was observed in the Elbow method curve. The Silhouette diagram, a plot of the number of clusters versus silhouette score exhibited a clear turning point at $k=15$, where the score reached the global maximum. Therefore, we grouped the figures into 15 clusters. 

\begin{figure}
     \centering
     \includegraphics[width = {0.7\textwidth}]{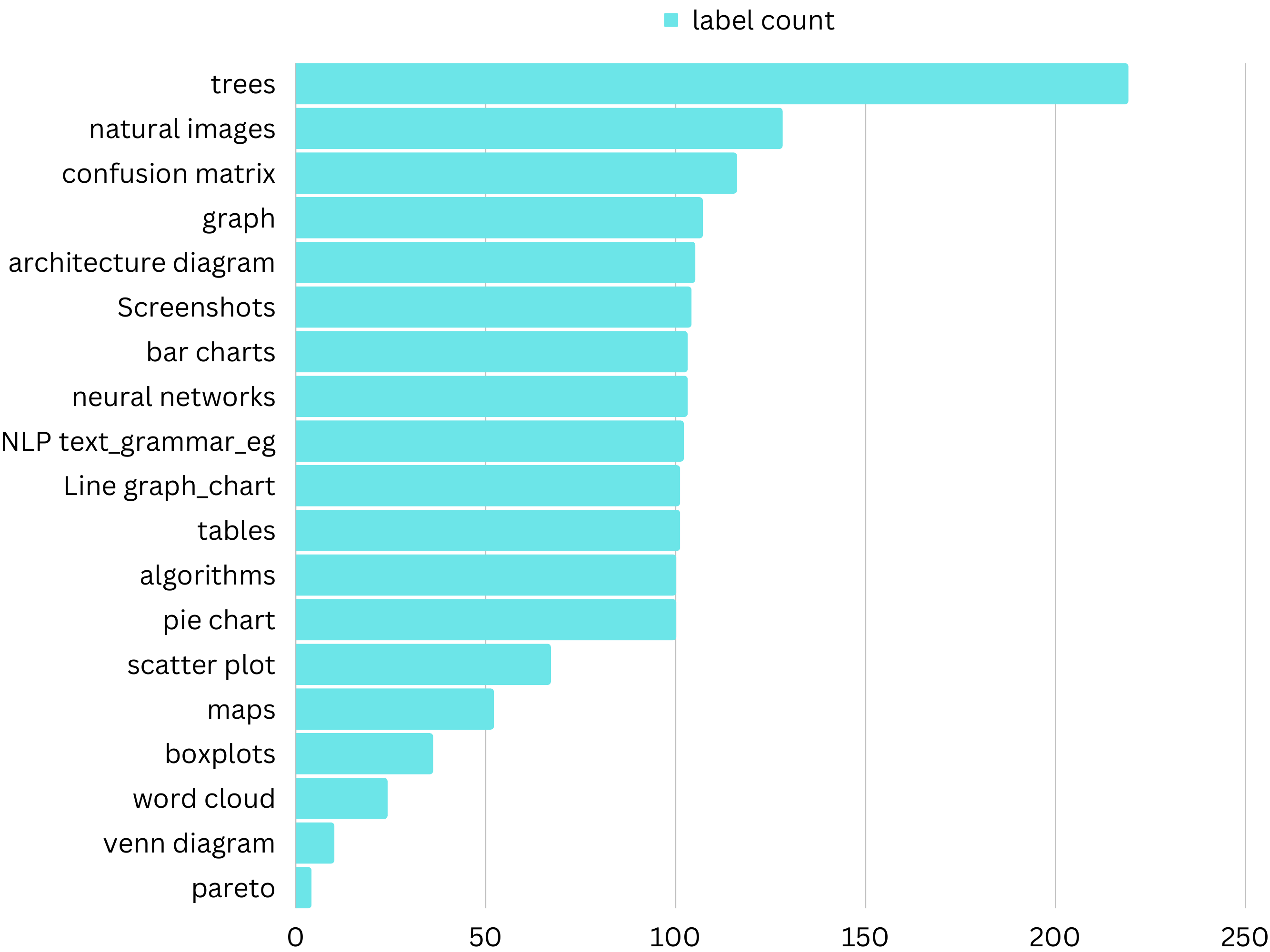}
     \caption{Figure class distribution in the \scifigp\ dataset.}
     \label{fig:pilotdist}
 \end{figure}

To validate the clustering results, 100 figures randomly sampled from each cluster were visually inspected. During the inspection, we identified three new figure types: \emph{word cloud}, \emph{pareto}, and \emph{venn diagram}. The \scifigp\ dataset was then built using all manually inspected figures. Two annotators manually labeled and inspected these clusters. The consensus rate was measured using Cohen's Kappa coefficient, which was $\kappa-0.78$ (substantial agreement) for the {\scifigp} dataset.  
For completeness, we added 100 randomly selected tables. Therefore, the \scifigp\ dataset contains a total of 1671 figures and tables labeled with 19 classes. The distribution of all classes is shown in Figure~\ref{fig:pilotdist}. 

\section{Supervised Scientific Figure Classification}
Based on the \scifigp\ dataset, we train supervised classifiers. The dataset was split into a training and a  test set (8:2 ratio). Three baseline models were investigated. Model~1 is a 3-Layer CNN, trained with a categorical cross-entropy loss function and the Adam optimizer. The model contains three typical convolutional layers, each followed by a max-pooling and a drop-out layer, and three fully-connected layers. The dimensions are reduced from $32\times32$ to $16\times16$ to $8\times8$. The last fully connected layer classifies the encoded vector into 19 classes. 
This classifier achieves an accuracy of 59\%. 

Model~2 was trained based on the VGG16 architecture ,except that the last three fully-connected layers in the original network were replaced by a long short-term memory layer, followed by dense layers for classification. This model achieved an accuracy of $\sim79\%$, 20\% higher than Model~1. 

Model~3 was the Vision Transformer (ViT) \cite{dosovitskiy2020image}, in which a figure was split into fixed-size patches. Each patch was then linearly embedded, supplemented by position embeddings. The resulting sequence of vectors was fed to a standard Transformer encoder.  The ViT model achieved the best performance, with 83\% accuracy. 


\section{Conclusion}
Based on the ACL Anthology papers, we designed a pipeline and used it to build a corpus of automatically labeled scientific figures with associated metadata and context information. This corpus, named \scifig, consists of $\approx250$k objects, of which about 42\% are figures and about 58\% are tables. We also built \scifigp, a subset of \scifig, consisting of 1671 scientific figures with 19 manually verified labels. Our dataset includes figures extracted from real-world data and contains more classes than existing datasets, e.g., DeepFigures and FigureQA. 

One limitation of our pipeline is that it used VGG16 pre-trained on ImageNet. In the future, we will improve  figure representation by retraining more sophisticated models, e.g., CoCa, \cite{yu2022coca}, on scientific figures. Another limitation was that determining the number of clusters required visual inspection. We will consider density-based methods  
to fully automate the clustering module. 

\bibliographystyle{unsrt}  
\bibliography{references}


\end{document}